\documentclass[letterpaper, 10 pt, conference]{ieeeconf}  

\IEEEoverridecommandlockouts                              

\usepackage{graphics} 
\usepackage{epsfig} 
\usepackage{amsmath} 
\usepackage{amssymb}  
\usepackage{tabularx}
\usepackage{booktabs}

\usepackage{fancyhdr}
\usepackage{xcolor}

\title{\LARGE \bf
Human-Robot collaboration in surgery: Advances and challenges towards autonomous surgical assistants}

\author{Jacinto Colan$^{1}$, Ana Davila$^{2}$, Yutaro Yamada$^{1}$ and Yasuhisa Hasegawa$^{2}$
\thanks{$^{1}$\quad
Department of Micro-Nano Mechanical Science and Engineering, Nagoya University, Furo-cho, Chikusa-ku, Nagoya, Aichi 464-8603, Japan}
\thanks{$^{2}$\quad
Institutes of Innovation for Future Society, Nagoya University, Furo-cho, Chikusa-ku, Nagoya, Aichi 464-8601, Japan}
\thanks{Correspondence: {\tt\small colan@robo.mein.nagoya-u.ac.jp}}
\thanks{
This work was supported in part by the Japan Science and Technology Agency (JST) CREST under Grant JPMJCR20D5, and in part by the Japan Society for the Promotion of Science (JSPS) Grants-in-Aid for Scientific Research (KAKENHI) under Grant 22K14221.}
}

\fancypagestyle{firstpage}{
  \fancyhf{}
  
  \fancyhead[L]{
    \vspace*{-25pt} 
    \parbox{\textwidth}{
      \textcolor{blue}{\small This work has been accepted at the 2025 IEEE International Conference on Robot and Human Interactive Communication (ROMAN) and submitted to the IEEE for possible publication. Copyright may be transferred without notice, after which this version may no longer be accessible.}
    }
  }
}

\begin{document}

\maketitle
\thispagestyle{firstpage}
\pagestyle{empty}

\begin{abstract}
Human-robot collaboration in surgery represents a significant area of research, driven by the increasing capability of autonomous robotic systems to assist surgeons in complex procedures. This systematic review examines the advancements and persistent challenges in the development of autonomous surgical robotic assistants (ASARs), focusing specifically on scenarios where robots provide meaningful and active support to human surgeons. Adhering to the PRISMA guidelines, a comprehensive literature search was conducted across the IEEE Xplore, Scopus, and Web of Science databases, resulting in the selection of 32 studies for detailed analysis. Two primary collaborative setups were identified: teleoperation-based assistance and direct hands-on interaction. The findings reveal a growing research emphasis on ASARs, with predominant applications currently in endoscope guidance, alongside emerging progress in autonomous tool manipulation. Several key challenges hinder wider adoption, including the alignment of robotic actions with human surgeon preferences, the necessity for procedural awareness within autonomous systems, the establishment of seamless human-robot information exchange, and the complexities of skill acquisition in shared workspaces. This review synthesizes current trends, identifies critical limitations, and outlines future research directions essential to improve the reliability, safety, and effectiveness of human-robot collaboration in surgical environments.
\end{abstract}

\section{INTRODUCTION}

Surgical robotics has substantially reshaped modern operative workflows; however, current systems operate primarily under direct teleoperated control, thereby limiting their potential as truly collaborative partners. Existing robotic systems exhibit a spectrum of autonomy, ranging from shared control approaches, where robots and surgeons jointly execute tasks aided by virtual fixtures or guidance systems \cite{yamada23task, colan21optimization}, to concepts for fully autonomous systems designed to independently perform discrete surgical subtasks with minimal human intervention \cite{saeidi22autonomous}. Both ends of this spectrum present distinct limitations: shared control often restricts the robot's autonomous decision-making capabilities and human control authority, while full autonomy raises safety concerns regarding the inherent variability and unpredictability of surgical environments, as well as the system's ability to achieve nuanced contextual understanding and comprehensive situational awareness \cite{attanasio21autonomy}.

\begin{figure}[t]
  \centering
  \includegraphics[width=0.75\linewidth]{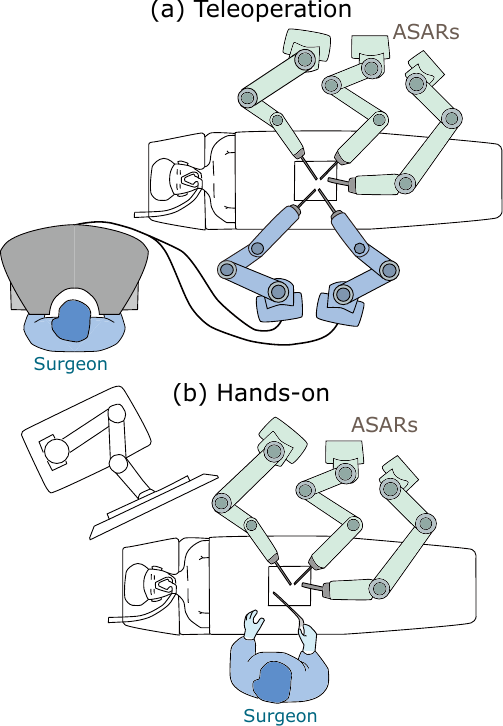}
  \caption{RMIS collaboration setups: (a) Teleoperation, where a lead surgeon operates primary instruments (blue) from a console while the gray autonomous surgical assistant robot (ASAR) assists. (b) Hands-on, with the surgeon manually operating alongside the gray ASAR at the sterile field.}
  \label{fig:1}
\vspace*{-5mm}
\end{figure}

In response to these challenges, human-robot collaboration (HRC) is an actively explored approach in surgical robotics. In HRC, autonomous systems are viewed as supportive partners working with surgeons, combining the strengths of human judgment and skill with robotic precision and stability. This approach often uses autonomous surgical assistant robots (ASARs) designed to perform specific support tasks, such as camera navigation or tissue hold, within robotic-assisted minimally invasive surgery (RMIS). The primary surgeon maintains overall control of the procedure. This collaboration can support solo surgery, allowing a single surgeon to manage procedures with robotic help. As shown in Figure~\ref{fig:1}, HRC in RMIS typically occurs in two main setups: (a) Teleoperation setups, where the lead surgeon operates instruments from a console while ASARs assist, and (b) Hands-on setups, where the surgeon works directly in the surgical area, in the same space as the ASARs. In both setups, ASARs work autonomously to help the primary surgeon.

Despite growing interest in this field, the existing literature lacks a comprehensive overview of advances, challenges, and collaboration details relevant to ASARs. Previous reviews have covered HRC generally or focused on specific technologies \cite{wan25review}. However, less attention has been paid to systematically analyzing the interaction details and human factors needed for safe and effective collaboration when robots provide autonomous assistance. This review addresses this gap by systematically examining studies on ASARs, focusing on collaboration challenges and human factors outcomes to provide a comprehensive overview of the current understanding of human-robot interaction in this context.

This paper seeks to:

\begin{itemize}
    \item Identify and describe different ways in which autonomy is implemented in robotic assistants for RMIS.
    \item Review the reported effectiveness of ASARs in various applications, focusing on human factors and collaborative performance.
    \item Discuss the specific challenges found in the development and implementation of these technologies.
\end{itemize}
\section{METHODOLOGY}
\subsection{Research question}
We defined the research question for this systematic review using the Population, Intervention, Comparators, and Outcomes (PICO) framework \cite{eriksen18impact}. This framework helps create a structured approach for identifying relevant studies and guides the selection criteria, as detailed in Table~\ref{tab:1}.

\begin{table}[bt] 
\caption{PICO framework outlining the research question components and inclusion/exclusion criteria\label{tab:1}}
\newcolumntype{C}{>{\centering\arraybackslash}X}
\begin{tabularx}{\linewidth}{p{0.15\linewidth} p{0.37\linewidth} p{0.37\linewidth}}
\toprule
\textbf{ }& \textbf{Inclusion criteria} & \textbf{Exclusion criteria}\\
\midrule
 Population & Studies on RMIS where an autonomous system (ASAR) assists the surgeon in surgical tasks. & Conventional MIS, teleoperated RMIS, or shared control systems without an autonomous assistant role.\\
 Intervention & Studies investigating autonomous robotic assistance in hands-on or teleoperation collaboration. &  Studies involving non-autonomous robotic systems or passive surgical tools. \\
 Comparators & Not applicable & - \\
 Outcomes &  Studies reporting on collaboration effectiveness, human factors, surgical task performance, or technical challenges. &Studies lacking empirical validation of their impact on surgical tasks or human factors.\\
\bottomrule
\end{tabularx}
\end{table}
\unskip

The research question for this review is: \textit{What are the advances and challenges of autonomous or semi-autonomous robotic systems assisting surgeons in RMIS within hands-on or teleoperation scenarios, based on empirically validated studies?}

This review aims to provide a clear overview of the current state of autonomous surgical assistants in RMIS, focusing on studies reporting experimental results. We explore progress and remaining challenges, concentrating on independent autonomous assistants rather than shared control methods or fully autonomous surgical pipelines. Furthermore, we focus on studies that included human factors assessment to understand effects related to the surgical team.

\subsection{Search methodology and systematic review}

\begin{figure}[t]
  \centering
  \includegraphics[width=0.9\linewidth]{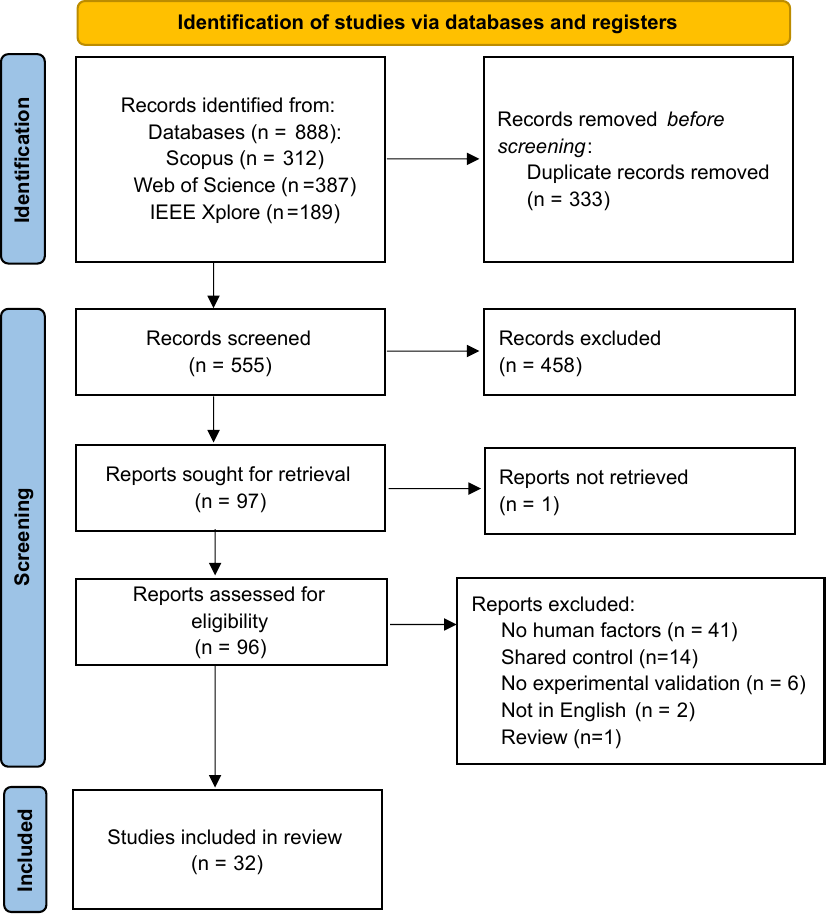}
  \caption{Diagram of the study selection process for this systematic review (following PRISMA).}
  \label{fig:2}
\vspace*{-5mm}
\end{figure}

This systematic review followed the guidelines in the Preferred Reporting Items for Systematic Reviews and Meta-Analyses (PRISMA) statement \cite{moher15preferred} to ensure a clear and repeatable process. Figure~\ref{fig:2} shows the study selection process. We searched three academic databases: Web of Science, Scopus, and IEEE Xplore Digital Library. The search covered publications from January 2015 to December 2025.  This timeframe was selected to include recent work on autonomous surgical assistant robots (ASARs). We used keywords related to robot-assisted minimally invasive surgery and human-robot collaboration. Table~\ref{tab2} shows the specific search terms used.

\begin{table}[bt] 
\caption{Search strategy\label{tab2}}
\newcolumntype{C}{>{\centering\arraybackslash}X}
\begin{tabularx}{\columnwidth}{CCC}
\toprule
\textbf{Search strategy}\\
\midrule
TITLE-ABS-KEY (
    robot* AND surg* AND ( "minimally invasive" OR laparoscopic OR endoscopic OR mis OR ramis OR rmis ) AND ( collaborat* OR assist* OR cooperat* OR help* ) AND autonom* ) ) 
AND PUBYEAR $>$ 2014 \\
\bottomrule
\end{tabularx}
\end{table}
\unskip

\subsection{Study Selection and Inclusion criteria}

We only included articles published in English. First, we removed duplicate records. During the screening phase, we reviewed titles and abstracts to exclude materials not peer-reviewed (brief communications, discussions, posters, project proposals) and articles not related to autonomous robotic assistance in RMIS. We also excluded surveys and review articles at this stage. We focused on studies about ASARs. We included only studies that reported results on human experience or performance in surgical tasks, evaluated through methods such as virtual simulations, phantom models, or \textit{ ex vivo} procedures. We excluded studies on conventional MIS, purely teleoperated RMIS, or shared control systems without a clear autonomous assistant role. Publications in which the full text was not accessible were also excluded.

Our search across the three databases resulted in 888 records (Scopus: 312, Web of Science: 387, IEEE Xplore: 189). After removing 333 duplicates, 555 articles remained for screening. Reviewing titles and abstracts led to the exclusion of 458 articles because they were not relevant (n=437) or were surveys/reviews (n=21). We could not access the full text for 1 record. This left 96 articles for full-text assessment. During the full-text review, we excluded studies for several reasons: 41 did not report on human factors; 14 focused on shared control without a specific autonomous assistant role; 6 lacked experimental validation; 2 were not in English; and 1 was identified as a review article.

\section{Results}

\subsection{Descriptive analysis of reviewed studies}

\begin{figure}[t]
  \centering
  \includegraphics[width=\linewidth]{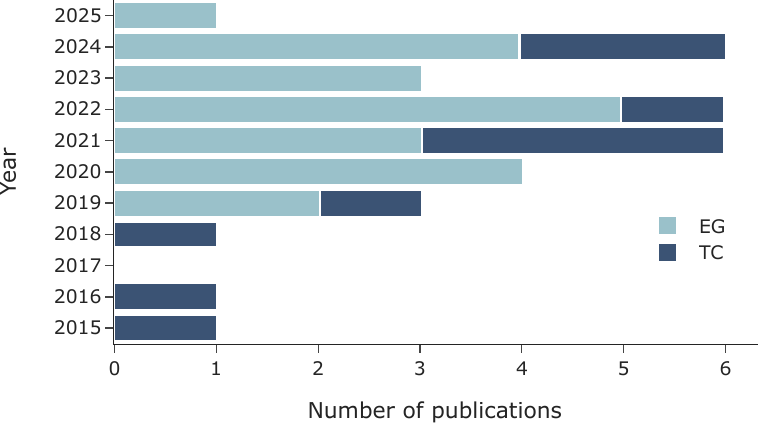}
  \caption{Number of publications per year (2015-2024) included in the review, categorized by primary assistance task. (EG: Endoscope guidance, TC: Tool control)}
  \label{fig:3}
\vspace*{-5mm}
\end{figure}

Figure~\ref{fig:3} shows the distribution of the included publications per year, categorized by the main assistance task studied. The analysis shows a notable increase in research on autonomous surgical assistance from 2018 onwards, increasing more rapidly from 2020-2024. The main assistance tasks were endoscope guidance (EG) and tool manipulation (TM). The stacked bars show the total count for each year; for instance, in 2021, six studies were included, composed of three focused on Tool Control (TC) and three on Endoscope Guidance (EG). Endoscope guidance was the focus of about 70\% of the reviewed studies. Tables~\ref{tab:3} (Teleoperation) and ~\ref{tab:4} (Hands-on) provide detailed summaries of the included studies, covering surgical tasks, assistant setup, autonomous capabilities, and reported human factors outcomes.

The reviewed studies show increasing variety in robotic platforms used, although the da Vinci Research Kit (dVRK) was the most commonly used platform (53\%). Most evaluations used phantom models (62\%), with simulation (19\%) and ex-vivo studies (19\%) making up the rest.

\begin{table*}[ht]
    \centering
    \renewcommand{\arraystretch}{1.2} 
    \caption{Summary of autonomous surgical assistance studies in teleoperated RMIS.}
    \scalebox{0.78}{
    \begin{tabular}{l c c p{1.4cm} p{1.4cm} p{3.4cm} p{9.6cm} c}
        \hline
        \textbf{Ref} & \textbf{Surgeon} & \textbf{Assist.} & \textbf{Robotic} & \textbf{Human} & \textbf{Autonomous} & \textbf{Human Factors Outcomes} & \textbf{Eval} \\ 
        & \textbf{Task} & \textbf{Task} & \textbf{Config} & \textbf{Interface} & \textbf{Capabilities} & & \\
        \hline
        \cite{shamaei15paced} & TM & TC & dVRK & dVRK & Paced shared-control with dominance factors for tissue manipulation & Improved performance with automated pulling vs. purely teleoperated operation. Reduced completion time with 100\% autonomous completion rate. Most cut paths remained within ±5mm boundaries. & P \\
        
        \cite{watanabe18single} & ST & TC & Pneumatic system & Phantom Premium & Force/rotation-triggered needle manipulation & Overall suturing time reduced by 20\%. Grasping/pulling phase reduced by 60\%. Significantly decreased variance in completion time compared to manual control. & P \\
        
        \cite{rivasblanco18transferring} & PP & EG & CISOBOT + UR3 & Phantom Omni & Learning-based camera guidance with trained behavioral model & Reduced surgeon workload by eliminating need for manual camera control. Appropriate views maintained throughout procedures with 78,712 data points recorded during 10-minute pick-and-place tasks. & P \\
                
        \cite{derossi19cognitive} & PP & TC & SARAS arm & dVRK & Action recognition with velocity modulation & Successful autonomous grasping and object placement with confidence levels generally above 50\%. Adaptive velocity modulation based on confidence thresholds. & P \\
        
        \cite{dacol20scan} & PP & EG & dVRK & dVRK & Instrument tracking-based endoscope navigation & Highest user confidence reported with autonomous control. Reduced cognitive workload (NASA-TLX). Better objective performance metrics related to surgical workflow optimization. & P \\
        
        \cite{mariani20experimental} & PP & EG & dVRK & dVRK & Training-optimized camera motion & Significantly higher time-accuracy metrics for users with autonomous camera navigation. Effective endoscope management imprinting for novice trainees (n=26). Validated effectiveness in early training stages. & S \\
        
        \cite{sun20visual} & PP & EG & Jaco2 & Omega 7 & Visual servoing with safety-RCM model & Reduced laparoscope motion frequency. Effective tracking performance validated in manipulation tasks. & P \\
        
        \cite{dacol21automating} & TM,ST & EG & dVRK & dVRK & Kinematic instrument tracking & Comparable procedure time with significantly reduced instruments-out-of-view time. SUS score 73.25 (above "good" threshold). Less experienced surgeons showed greater benefit and higher usability ratings. & E \\

        \cite{barragan21sachets} & TM & BS & dVRK & dVRK & FCN-based blood detection and suction & Significantly reduced workload demands (NASA-TLX) in autonomous mode. Decreased end-effector velocity, number of clutching events, and total clutching time. & P \\
        
        \cite{scheikl21cooperative} & TM & TC & Simulated robot & Xbox controllers & Multi-agent reinforcement learning & Hybrid human-AI teams achieved 100\% success rate with lower collision rates when humans controlled grippers and AI controlled cautery. NASA-TLX showed cauter control perceived as twice as demanding as gripper control. & S \\

        \cite{derossi21first} & PP & TC & SARAS arms & dVRK & Multi-modal cognitive system with action segmentation and safety-bounded control & Effective coordination in surgeon-robot handover tasks. Successful execution of complementary actions based on recognized phases. Demonstrated feasible cooperative manipulation & P \\
        
        \cite{pasini22virtual} & ST & EG & dVRK & dVRK & Situation-aware camera motion & Reduced mental and physical workload for surgeons. Shorter surgery times validated in user study. Improved surgical workflow through enhanced camera control system. & S \\
        
        \cite{elazzazi22natural} & - & EG & dVRK & dVRK & Natural language camera control & Reduced need for manual camera manipulation. Audio feedback confirmed recognized commands. Successful execution of voice commands with comparative recognition accuracy between online and offline systems. & S,P \\
        
        \cite{pasini23grace} & ST & EG & dVRK & dVRK & Gesture recognition for camera enhancement & Statistically significant reduction in NASA-TLX workload and motion sickness  compared to standard control. Improved workflow by enabling simultaneous instrument and camera control. & P \\
        
        \cite{gao23savanet} & PP,ST & EG & Sim dVRK & Not mentioned & Visual attention prediction for endoscope control & Maintained important surgical targets in field of view based on predicted surgeon attention points. Over 91\% accuracy in surgical action prediction during dynamic scenes. Effective handling of surgical target tracking & S \\
        
        \cite{chen24toward} & PP & EG & dVRK & dVRK & Context-aware autonomous navigation & Execution time reduced by 24.67\% compared to standard setup. Reduced workload demonstrated in 10-participant user study. Maintained good field of view through autonomous phase-based mode switching. & P \\
        
        \cite{dong24intelligent} & PP & EG & Sim dVRK & Geomagic Touch & NLP, CV, and procedure perception models & Higher success rates with autonomous control. Reduced task completion time in both autonomous and semi-autonomous modes. Improved human-robot communication with procedure logging. & S \\
        
        \cite{strohmeyer24semi} & TM & TC & dVRK & dVRK & Model-free tissue tensioning during cutting & Successful cutting task performance across 25 experiments on silicone phantoms without requiring prior tissue property knowledge. Effective autonomous tissue tensioning during surgeon cutting. & P \\
        
        \cite{sun25sgr} & PP & EG & SGR-AutoLap & SGR-AutoLap & Gesture recognition-based camera control & Reduced task execution time compared to manual interruption mode. Improved surgeon comfort scores (0-10 scale). Enhanced efficiency without workflow interruption using gesture recognition framework. & P \\ 
        \hline
        \multicolumn{8}{l}{TM: Tissue manipulation, ST: Suturing, PP: Pick-and-place, TC: Tool control, EG: Endoscope guidance, BS: Blood suction}\\
        \multicolumn{8}{l}{Evaluation: S = Simulation, P = Phantom, E = \textit{Ex-vivo} study}
    \end{tabular}}
    \label{tab:3}
\end{table*}

\begin{table*}[ht]
    \centering
    \renewcommand{\arraystretch}{1.2} 
    \caption{Summary of autonomous surgical assistance studies in hands-on RMIS.}
    \scalebox{0.78}{
    \begin{tabular}{l c c p{1.2cm} p{5cm} p{10.5cm} c}
        \hline
        \textbf{Ref} & \textbf{Surgeon} & \textbf{Assist.} & \textbf{Robotic} & \textbf{Autonomous} & \textbf{Human Factors Outcomes} & \textbf{Eval} \\ 
        & \textbf{Task} & \textbf{Task} & \textbf{Config} & \textbf{Capabilities} & & \\
        \hline
        
        \cite{bauzano16collaborative} & ST & TC & CISOBOT & Task recognition system with autonomous actuation based on surgical workflow model & Surgeon gesture recognition accuracy of 80\% (reducing required voice commands). Comparable completion times between robot assistance and human assistance. Validation with 15 trials comparing human vs. robot assistance for complete suture procedures. & P \\
        
        \cite{ma19autonomous} & - & EG & dVRK & Autonomous flexible endoscope with visual servoing and optimal control minimizing motion and space occupation & Reduce internal and external space compared to rigid endoscopes. user study (n=10) validated wider field of view and safer operation; improved surgical efficiency through optimized space management & P,E \\
        
        \cite{song20robotic} & TM & EG & dVRK & Robotic flexible endoscope (RFE) with enhanced field of view and dexterity for minimizing space occupation & Operation time with RFE comparable to manual and significantly shorter than rigid endoscopes. NASA-TLX showed lower workload ratings. Improved performance due to reduced instrument interference. & E \\
        
        \cite{wagner21learning} & TM & EG & LWR 4 + VIKY EP & Learning-based camera guidance that adapts to surgeon's preferences through experience using a knowledge-based approach & Reduced operation duration across learning phases. Decreased camera interactions indicating improved surgeon satisfaction. Improved camera guidance quality ratings. & P \\
        
        \cite{sandoval21towards} & PP,ST & EG & Franka Emika & Autonomous tracking of surgical instruments using visual servoing & Significantly reduced angular workspace required by surgeon compared to medical assistant. Improved ergonomics and workspace management. Reduced surgeon physical workload. & P \\
        
        \cite{peng22endoscope} & - & EG & myCobot Pro 600 & Dual-mode system with autonomous FOV tracking and speech-based control with voice commands & Fast/accurate instrument tip tracking with reduced hand-eye incoordination. Improved field-of-view stability. Dual control modality provided flexibility in surgical workflow. Reduced surgeon attention division. & S,P \\
        
        \cite{huang22surgeon} & TM & EG & dVRK & Intelligent endoscope control system with tissue tracking and occlusion handling & Successfully completed tissue resection task in 1.5 minutes with 10mm-2mm tissue samples. System maintained stable imaging during tissue occlusions. Continuous field-of-view optimization during dynamic surgical manipulation. & E \\
        
        \cite{gruijthuijsen22robotic} & PP & EG & Virtuose 6D & Robotic endoscope holder with autonomous guidance for minimally invasive surgery & 40 trials completed with average completion time of 172s. Observable learning curve for both novices and surgeons. Users successfully performed benchmark surgical tasks with autonomous endoscope guidance within allocated time constraints. & P,E \\
        
        \cite{zini22cognitive} & TM & TC & UR5 & Action recognition module with supervisory controller for autonomous assistance & Successful task synchronization between surgeon and robot with proper action sequencing. Accurate recognition and execution of complementary surgical tasks. Improved collaborative workflow through cognitive architecture. & P \\
        
        \cite{fozilov23endoscope} & TM & EG & Gen3 & Multi-tool detection with ROI estimation for dynamic FOV adjustment & Optimized endoscope following rate (FR) and reduced total manipulation time. System maintained tracking error below 50 pixel threshold. NASA-TLX assessment demonstrated reduced cognitive load and fatigue compared to manual control. & P \\
        
        \cite{huang24visual} & TM & EG & LBR Med 7 & 4-DOF visual servoing with neural network optimization and predefined-time convergence for tracking with noise rejection & Completed lung tissue resection in approximately 25 seconds with five tissue samples successfully resected. System maintained stability with RCM position error less than 2.5 cm. Smooth tracking during dynamic tissue manipulation. & E \\
        
        \cite{ergen24design} & TM & EG & Stewart platform & Autonomous endoscope navigation along predetermined surgical routes with route recognition & Reduced system setup and surgery time. Diminished impact of operator experience compared to manual mode. Enhanced ergonomics and improved operating room workflow efficiency through autonomous navigation. & E \\
        
        \cite{liu24latent} & TM & TC & Gen3 & Latent Regression-Model Predictive Control (LR-MPC) for autonomous tissue triangulation & LR-MPC achieved performance comparable to human assistant in tissue triangulation tasks across various tissue sizes and materials. Consistent technique execution across different tissue properties. Improved procedural standardization. & P \\
        \hline
        \multicolumn{7}{l}{PP: Pick-and-place, TM: Tissue manipulation, ST: Suturing, EG: Endoscope guidance, TC: Tool control}\\
        \multicolumn{7}{l}{Evaluation: S = Simulation, P = Phantom, E = \textit{Ex-vivo} study}
    \end{tabular}}
    \label{tab:4}
\end{table*}

\subsection{Assistance tasks}

\subsubsection{Endoscope guidance (EG)}
Endoscope guidance is the most frequent application found in the reviewed studies (69\%), likely due to the cognitive and physical effort often needed for manual camera control during surgery. These systems aim to provide optimal visualization, allowing surgeons to focus more on the primary surgical tasks.

Approaches in this area have changed during the review period. Early systems mainly used instrument tracking \cite{dacol20scan, ma19autonomous, sandoval21towards, dacol21automating, gruijthuijsen22robotic, huang22surgeon, song20robotic}, with advances in computer vision models allowing real-time identification and tracking of instruments or tissues \cite{fozilov23endoscope}.
Recent work has focused on three main areas. First, context-aware systems have emerged that adapt camera positioning based on surgical phase recognition \cite{chen24toward}, moving beyond simple following to anticipatory positioning. Second, natural interaction methods including gesture recognition \cite{pasini22virtual, pasini23grace, sun25sgr} and voice commands \cite{elazzazi22natural, peng22endoscope, dong24intelligent} have been integrated to provide intuitive control without disrupting workflow. Third, learning-based approaches have enabled personalization to surgeon preferences through demonstration \cite{rivasblanco18transferring, wagner21learning}, creating more responsive and adaptable assistance.

Reported human factors outcomes generally show benefits. Workload assessments often reported reductions in cognitive and physical demands \cite{pasini23grace, dacol20scan}, helping surgeons dedicate more attention to the main surgical tasks. Efficiency improved in several studies, with shorter task times or fewer workflow interruptions \cite{chen24toward, dacol21automating}. Ergonomic benefits were often reported, such as reduced physical strain or better workspace management compared to manual camera holding \cite{sandoval21towards, ergen24design}. User satisfaction was generally positive, with surgeons in some studies feeling more confident and comfortable when using autonomous systems adapted to their needs \cite{wagner21learning, sun25sgr}. These improvements were sometimes greater for less experienced surgeons \cite{dacol21automating}, suggesting these systems could be useful for training.

\subsubsection{Tool manipulation (TC)}
Tool manipulation systems (31\% of the reviewed studies) are a more complex area of autonomous surgical assistance. This category includes tasks such as tissue manipulation (pulling, tensioning), suturing assistance, and specialized tasks such as automated blood suction. A key aspect of tool manipulation assistance is the need for close coordination between the autonomous system and the surgeon working in the same space. This requires advanced system abilities to perceive the situation, predict the surgeon's intention, and help appropriately without interruption.

Applications in this category show growing complexity. Tissue manipulation is the most frequent task, with different methods used to help with pulling tissue \cite{shamaei15paced}, providing tension during cutting \cite{strohmeyer24semi}, or achieving triangulation \cite{liu24latent}. Suturing assistance is another key area, where systems might automate parts of the task like needle handling \cite{watanabe18single, bauzano16collaborative} while the surgeon controls critical steps. Specialized applications include automated blood suction \cite{barragan21sachets} and handling objects \cite{derossi19cognitive, derossi21first}. These systems use different coordination methods, such as reinforcement learning for shared tasks \cite{scheikl21cooperative} or recognizing workflow steps to provide timely help \cite{zini22cognitive}.

Reported human factors outcomes generally show benefits here as well. Performance improvements reported include lower task completion times; for example, one study reported substantial decreases in suturing time \cite{watanabe18single}. Workload assessments showed benefits, especially for demanding secondary tasks like managing blood suction \cite{barragan21sachets}. Quality measures in some studies suggested precision was maintained or improved compared to manual assistance \cite{shamaei15paced}, with one study reporting 100\% success rates for a task completed by hybrid human-robot teams \cite{scheikl21cooperative}.

\section{Discussion}
\label{sec:4}

Despite substantial research efforts in autonomous surgical assistance, most technologies remain in pre-clinical development. This section discusses the key challenges identified in our review that represent the primary barriers to widespread clinical translation and adoption. 

\subsection{Remaining challenges}

\begin{table}[bt] 
\caption{Challenges associated to autonomous surgical assistants \label{tab:5}}
\newcolumntype{C}{>{\centering\arraybackslash}X}
\begin{tabularx}{\linewidth}{p{0.3\linewidth} p{0.59\linewidth}}
\toprule
\textbf{Challenge}& \textbf{Refs.} \\
\midrule
 Preference alignment & \cite{rivasblanco18transferring}, \cite{ma19autonomous}, \cite{dacol20scan}, \cite{mariani20experimental}, \cite{sun20visual}, \cite{song20robotic},  \cite{sandoval21towards}, \cite{dacol21automating}, \cite{scheikl21cooperative}, \cite{pasini22virtual}, \cite{peng22endoscope}, \cite{huang22surgeon}, \cite{gruijthuijsen22robotic}, \cite{fozilov23endoscope}, \cite{pasini23grace}, \cite{gao23savanet}, \cite{chen24toward}, \cite{huang24visual}, \cite{ergen24design}, \cite{sun25sgr}\\
 
 Procedural awareness & \cite{shamaei15paced}, \cite{bauzano16collaborative}, \cite{watanabe18single}, \cite{derossi19cognitive}, \cite{wagner21learning}, \cite{barragan21sachets},  \cite{derossi21first}, \cite{pasini22virtual}, \cite{zini22cognitive}, \cite{pasini23grace}, \cite{dong24intelligent}, \cite{chen24toward}, \cite{strohmeyer24semi}, \cite{sun25sgr}  \\
 
  Skill acquisition & \cite{rivasblanco18transferring}, \cite{wagner21learning}, \cite{scheikl21cooperative}, \cite{strohmeyer24semi}, \cite{liu24latent} \\
  
  Information exchange & \cite{bauzano16collaborative}, \cite{elazzazi22natural}, \cite{peng22endoscope}, \cite{dong24intelligent} \\

\bottomrule
\end{tabularx}
\end{table}
\unskip

\subsubsection{Human preference alignment}
Human preference alignment is identified as a common challenge (20 studies), particularly in endoscope guidance systems. Surgical assistance systems must account for individual surgeon preferences, which vary based on training background, experience level, and procedural approach.

Several studies have explored approaches to address this variability. Wagner et al. \cite{wagner21learning} implemented learning-based camera guidance that adapts to surgeon preferences, showing improved camera guidance ratings following adaptation periods. Pasini et al. \cite{pasini23grace} utilized gesture recognition to allow surgeons to communicate camera positioning preferences during procedures. Huang et al. \cite{huang22surgeon} defined instrument tracking preferences by analyzing operation videos and consulting surgeons.

Experience level appears to influence system acceptance and benefit. Dacol et al. \cite{dacol21automating} observed that less experienced surgeons demonstrated greater benefit and higher usability ratings for autonomous systems compared to more experienced surgeons, indicating that preference alignment may need to account for experience level.

The research suggests that effective autonomous surgical assistants require mechanisms to adapt to individual surgeon preferences while maintaining procedural standards. This represents a technical challenge that involves balancing personalization with consistent performance across different users. Current approaches include demonstration-based learning \cite{rivasblanco18transferring}, adaptive frameworks that learn from experience \cite{wagner21learning}, and systems that provide standardization while accommodating individual variations \cite{liu24latent, ergen24design}. However, these methods often demand extensive user-specific tuning or long adaptation, impractical in clinical settings and struggling to generalize across diverse surgical techniques, thus lacking one-shot adaptability.

\subsubsection{Surgical Procedural awareness}
Procedural awareness is the capability to comprehend the surgical workflow and context. It has been identified as a challenge in 14 studies. In contrast to industrial automation, which follows fixed sequences, surgical procedures are dynamic and involve various granular levels \cite{yamada24multimodal}. Different phases necessitate distinct types of assistance and require an understanding of the overall procedural objectives.

Research approaches have evolved from simplistic rule-based methods to more comprehensive understanding frameworks. Chen et al. \cite{chen24toward} implemented context-aware navigation that automatically switches between tracking modes based on surgical phase recognition, reducing execution time compared to standard setups. Similarly, Sun et al. \cite{sun25sgr} used surgical gesture recognition to provide optimal views based on the current procedural step. Bauzano et al. \cite{bauzano16collaborative} incorporated task recognition systems with surgical workflow models to guide appropriate assistance.

Several studies have addressed procedural awareness through integrated perception systems. Derossi et al. \cite{derossi21first} developed multi-modal cognitive systems for action segmentation and supervisory control, enabling adaptation of assistance based on recognized surgical actions. Zini et al. \cite{zini22cognitive} demonstrated successful task synchronization between surgeon and robot through cognitive architectures with action recognition capabilities. Gao et al. \cite{gao23savanet} achieved a high precision in surgical action prediction to maintain the desired objectives in the field of view.

Current limitations include reliance on predefined procedural models, which may not account for unexpected situations or procedural variations. The integration of machine learning approaches for phase recognition and need anticipation presents a promising direction for overcoming these limitations. Dong et al. \cite{dong24intelligent} combined procedure perception models with natural language processing and computer vision to improve success rates, demonstrating the potential benefits of comprehensive procedural understanding in autonomous surgical systems.

\subsubsection{Collaborative skill acquisition}
The challenge of collaborative skill acquisition, how autonomous systems learn tasks requiring coordination between multiple agents, was identified in five studies. Although datasets exist for individual surgical skills, collaborative tasks present a fundamental data acquisition problem: how to learn behaviors that depend on interaction between multiple parties.

Liu et al. \cite{liu24latent} addressed this by collecting human-performed tissue triangulation demonstrations, achieve performance comparable to human assistants across various tissue types. Strohmeyer et al. \cite{strohmeyer24semi} developed a model-free approach that adapts tensioning behavior based on real-time surgeon interaction rather than requiring extensive prior demonstrations. Rivas-Blanco et al. \cite{rivasblanco18transferring} collected a large number of data points from human demonstrations to create a behavioral model for camera control. Scheikl et al. \cite{scheikl21cooperative} explored multi-agent reinforcement learning where simulated agents learn collaborative policies without human demonstration data, but highlighted difficulties in accurately modeling human behavior in these simulations.

The fundamental challenge remains: collaborative surgical tasks require understanding and predicting human actions, but there is no comprehensive way to simulate the full range of potential human behaviors for training. This creates a significant barrier to developing systems that can effectively learn collaborative surgical skills without extensive human-human demonstration data.

\subsubsection{Human-robot information exchange}
Human-robot communication was identified as a challenge in four studies, focusing on the need for intuitive, reliable information exchange between surgeons and autonomous systems. Recent advancements in natural language interfaces for surgical robots \cite{elazzazi22natural}, \cite{davila24voice}, \cite{pandya23chatgpt} underscore the growing importance of accurate language interpretation in surgical contexts.

Elazzazi et al. \cite{elazzazi22natural} and Peng et al. \cite{peng22endoscope} explored natural language interfaces for autonomous control, reducing the need for manual camera manipulation while providing feedback mechanisms for command confirmation. These approaches show promise for explicit communication, but face challenges in noisy operating room environments and can fail to capture the non-verbal cues surgeons use implicitly. Bauzano et al. \cite{bauzano16collaborative} combined surgeon gesture recognition with voice commands, maintaining comparable completion times between robot and human assistance. This implicit communication approach may reduce cognitive load compared to explicit command systems. Dong et al. \cite{dong24intelligent} integrated multiple communication modalities (NLP, CV, procedure perception) to improve success rates and task completion time, suggesting that multi-modal communication approaches may provide the most robust solution for surgical settings.

Future systems will likely need to combine explicit and implicit communication channels, with contextual awareness to interpret surgeon intent from minimal cues. The development of standardized communication protocols for surgical robots could facilitate interoperability and reduce the training burden between platforms.

\section{CONCLUSIONS}
\label{sec:5}

This systematic review on autonomous surgical assistants reveals a rapidly evolving field poised to transform surgical practice through systems that reduce workload and enhance procedural outcomes. Four critical challenges impede clinical adoption: human preference alignment (adapting to individual surgeon styles), procedural awareness (contextual understanding of workflows), collaborative skill acquisition (robots learning interactive tasks), and human-robot communication (intuitive information exchange). Future research should prioritize personalization frameworks, contextual understanding, collaborative data collection with data-efficient learning approaches, and multimodal interfaces with standardized evaluation metrics. Recent advances in large multimodal and foundation models offer promising potential for more intuitive interfaces, addressing key challenges. The evolution of autonomous surgical assistants represents a fundamental shift from tools operated by surgeons to intelligent collaborators that enhance human capabilities while preserving the critical judgment characteristic of expert surgical practice.

\addtolength{\textheight}{-0cm}   







\bibliographystyle{IEEEtran}
\bibliography{biblio}

\end{document}